%% file: 0_main.tex
\documentclass[10pt,twocolumn,letterpaper]{article}

\usepackage{iccv}

\input{config}

\iccvfinalcopy 

\def\iccvPaperID{9974} 

\ificcvfinal\fi

\begin{document}

\title{VLGrammar: Grounded Grammar Induction of Vision and Language}

\author{Yining Hong, Qing Li, Song-Chun Zhu, Siyuan Huang\\
University of California, Los Angeles, USA\\
{\tt\small yininghong@cs.ucla.edu, liqing@ucla.edu, sczhu@stat.ucla.edu, huangsiyuan@ucla.edu}
}

\maketitle
\ificcvfinal\thispagestyle{empty}\fi

\begin{abstract}
Cognitive grammar suggests that the acquisition of language grammar is grounded within visual structures. While grammar is an essential representation of natural language, it also exists ubiquitously in vision to represent the hierarchical part-whole structure. In this work, we study grounded grammar induction of vision and language in a joint learning framework. Specifically, we present VLGrammar, a method that uses compound probabilistic context-free grammars (compound PCFGs) to induce the language grammar and the image grammar simultaneously. We propose a novel contrastive learning framework to guide the joint learning of both modules. To provide a benchmark for the grounded grammar induction task, we collect a large-scale dataset, \textsc{PartIt}, which contains human-written sentences that describe part-level semantics for 3D objects. Experiments on the \textsc{PartIt} dataset show that VLGrammar outperforms all baselines in image grammar induction and language grammar induction. The learned VLGrammar naturally benefits related downstream tasks. Specifically, it improves the image unsupervised clustering accuracy by 30\%, and performs well in image retrieval and text retrieval. Notably, the induced grammar shows superior generalizability by easily generalizing to unseen categories.
\end{abstract}

\input{1_intro}
\input{2_related_work}
\input{3_dataset}
\input{4_method}
\input{5_experiment}

\input{6_conclusion}

{\small
\bibliographystyle{ieee_fullname}
\bibliography{reference}
}

\end{document}

%% file: config.tex
\usepackage{times}
\usepackage{graphicx}
\usepackage{amsmath,amssymb,amsthm,mathabx}
\usepackage{booktabs}
\usepackage{acronym}
\usepackage{enumitem}
\usepackage[pagebackref=true,breaklinks=true,colorlinks,bookmarks=false]{hyperref}
\usepackage{balance}
\usepackage{setspace}
\usepackage[skip=3pt,font=small]{subcaption}
\usepackage[skip=3pt,font=small]{caption}
\usepackage[dvipsnames,table]{xcolor}
\usepackage[capitalise]{cleveref}
\usepackage{tabularx,multirow,array,makecell}
\usepackage{pifont}
\usepackage{bbm}
\usepackage[capitalise]{cleveref}

\makeatletter
\renewcommand{\paragraph}{%
  \@startsection{paragraph}{4}%
  {\z@}{0ex \@plus 0ex \@minus 0ex}{-1em}%
  {\hskip\parindent\normalfont\normalsize\bfseries}%
}
\makeatother

\graphicspath{{figures/}}

\crefname{algocf}{Alg.}{Algs.}
\Crefname{algocf}{Algorithm}{Algorithm}

\definecolor{gblue}{HTML}{4285F4}
\definecolor{gred}{HTML}{DB4437}

\acrodef{amt}[AMT]{Amazon Mechanic Turk}

%% file: 1_intro.tex
\section{Introduction}

\begin{figure}[t!]
    \centering
    \includegraphics[width=\linewidth]{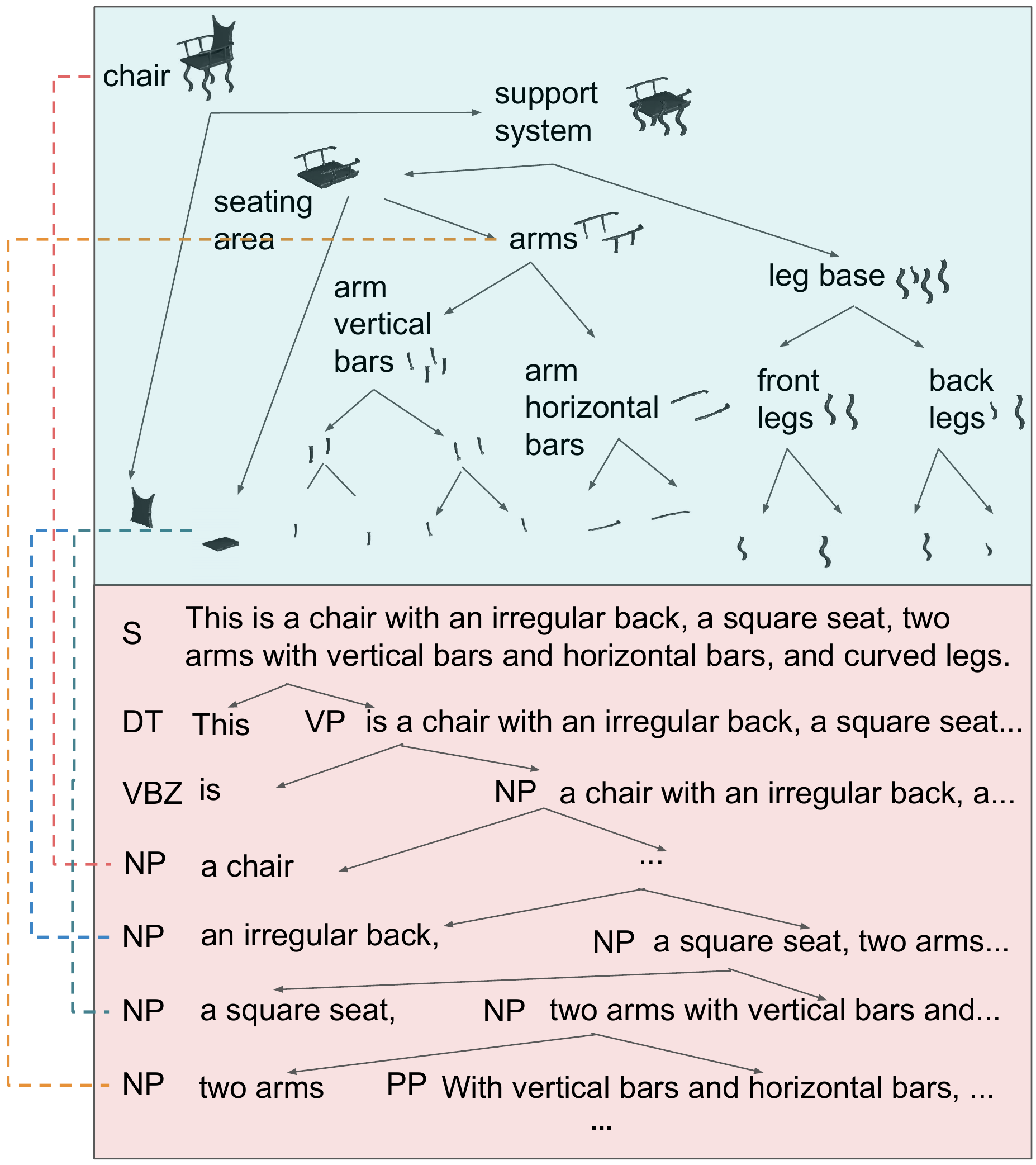}
    \caption{\textbf{An example of a sentence parse tree aligned with an image parse tree.} The arrow lines represent production rules of the image grammar and the language grammar. The dashed lines represent alignment between the constituents of two modalities.}
    \label{fig:intro}
\end{figure}

Natural and man-made dynamical systems tend to have a nested multi-scale organization, which might be a general property of all physical and biological systems. According to \cite{simon1973hierarchy}, building complex stable systems requires the re-use of stable sub-systems that can be assembled to build larger systems. Therefore, exploring the  low-dimensional structures in sensory data is critical for understanding the world and helping the design, interpretation, and generalization of artificial intelligent systems. Similarly, inducing the underlying structures and grammars from raw sensory inputs, \eg, vision and language~\cite{Fu1968SyntacticPR, You1978SyntacticSR,Solmon2003TheEO, Klein2002AGC, Simone2018GuidingNP, Cohen2008TheSL, Spitkovsky2010ViterbiTI,zhu2006stochastic,Wu2010ANS,Sabour2017DynamicRB, Hinton2021HowTR,li2020closed,huang2018holistic}, has been a long-standing challenge in the field of artificial intelligence (AI).

With the development of unsupervised learning techniques, the unsupervised grammar induction for natural language \cite{Shen2018NeuralLM, Shen2019OrderedNI, Kim2019UnsupervisedRN, Kim2019CompoundPC}  has recently made satisfying progress.  These works formulate the grammar induction of language as a self-contained system that relies solely on textual corpora. Following this trend, \cite{Shi2019VisuallyGN, Zhao2020VisuallyGC} propose the visually grounded grammar induction. They empirically show that if the constituents in a sentence's parse tree are well aligned with the image that the sentence describes, the induced grammar will be more accurate. 

Visually grounded grammar induction takes one step further towards \textit{cognitive grammar} \cite{Langacker1983FoundationsOC, Langacker1986AnIT}, a concept from linguistic theory. Cognitive grammar argues that it is pointless to analyze grammatical units without reference to their semantics,  which is grounded and structured by patterns of perception, such as vision. However, previous works ground all the constituents of a sentence with the embedding of a single image~\cite{Shi2019VisuallyGN, Zhao2020VisuallyGC}. They focus on aligning the image feature to language grammar but miss the hierarchical structures in the image. This is inconsistent with cognitive grammar's notion that a constituent's semantic value does not reside in one individual image base, but rather in the relationship between the substructure and the base.

Part-whole relationships are crucial in semantic structures \cite{Johnson1987TheBI}. For example, the constituent ``two arms" in \cref{fig:intro} does not simply refer to a chair, but instead refers to the chair's arms. Thus, it is necessary to align the language grammar with the hierarchical structures in physical objects. As shown in \cref{fig:intro}, a visual object can be parsed into parts with hierarchical structures, and constituents that describe parts of an object can be naturally grounded with the parts at different hierarchies. 


While the study of the hierarchical structure of images has a long history \cite{Fu1968SyntacticPR, You1978SyntacticSR, Sabour2017DynamicRB, Hinton2021HowTR, Han2009BottomUpTopDownIP, Tu2003ImagePU, zhu2006stochastic}, the structure is mainly pre-defined by human and static across images. Therefore, challenges remain as: (1) how to represent flexible part-whole hierarchies that vary with images using an identical network \cite{Hinton2021HowTR}, and (2) how to learn structure automatically without pre-defined templates. One possible way is to learn the image grammar that parses an object into parts. Instead of allocating neurons to represent nodes in the parse graph, we can use neurons to represent grammar rules. The grammar rules are general for all the images and can be recursively re-used to handle arbitrarily complicated objects (\eg, a chair can have an arbitrary number of legs).

Inspired by the above ideas, we present VLGrammar, a framework that jointly learns image and language grammar. Specifically, we use compound probabilistic context-free grammars (compound PCFGs), which parameterize a PCFG's rule probabilities with neural networks and relax the context-free constraints with a latent compound variable. To achieve grounded learning, we calculate an alignment score between the image parse tree and the language parse tree, and use a contrastive loss to learn the compound PCFGs for both image and language jointly.

To obtain data that contains multi-modal part-whole information for learning grounded grammars, we collect a large-scale dataset, \textsc{PartIt}, which contains 10,613 manually annotated descriptive sentences paired with the images of objects and parts. The sentences collected via \acf{amt} describe the detailed object and part semantics for 3D objects.

Experiments on the proposed \textsc{PartIt} dataset show that our proposed VLGrammar outperforms all baselines in both image grammar induction and language grammar induction. Moreover, it naturally benefits related downstream tasks, for example, improving the accuracy of unsupervised part clustering from $\sim$40\% to $\sim$70\%, and achieving better performance in the image-text retrieval tasks. Our image grammar trained on \texttt{chair} and \texttt{table} can be easily generalized to unseen categories such as \texttt{bed} and \texttt{bag}. Qualitative studies also show that our method is capable of predicting part-whole hierarchies and recursive structures of objects, as well as constituency parsing of sentences.

Our contributions can be summarized as follows:
\begin{itemize}[nosep]
    \item To benchmark the grounded grammar induction problem, we collect a large-scale dataset, \textsc{PartIt}, which contains human-written sentences describing both object-level and part-level semantics for 3D objects.
    \item We propose VLGrammar, which utilizes compound PCFGs to induce grounded grammars for both vision and language by enforcing the cross-modal alignment.
    \item We conduct extensive experiments on the \textsc{PartIt} dataset. The results demonstrate the superiority of the proposed VLGrammar on the grammar induction and downstream tasks, such as unsupervised part clustering and image-text retrieval. VLGrammar also shows great generalization ability on unseen object categories.
\end{itemize}

%% file: 2_related_work.tex
\section{Related Work}
\subsection{Grammar Induction of Language}
Grammar induction has a long history in natural language processing \cite{Solmon2003TheEO, Klein2002AGC, Simone2018GuidingNP, Cohen2008TheSL, Spitkovsky2010ViterbiTI}. 
Recently, researchers focus on using neural networks to induce parse trees solely from sentences \cite{Shen2019OrderedNI, Shen2018NeuralLM, Drozdov2019UnsupervisedLT, Kim2019UnsupervisedRN, Kim2019CompoundPC}.

These approaches mostly suggest that language is an autonomous system that does not rely on perceptions and semantics. This notion departs from the cognitive grammar concept in linguistic theory which emphasizes the role of semantic structure in grammar induction.  To address this issue, visually-grounded grammar induction is proposed  \cite{Shi2019VisuallyGN, Zhao2020VisuallyGC}. However, they use a single image to estimate the \textit{concreteness} of language spans (\ie, to define concrete words as those referring to perception) \cite{Turney2011LiteralAM, Kiela2014ImprovingMR}, which is insufficient to represent the full semantic structure. Given that a language constituent is typically associated with a specific part of the image, we propose to align a constituent with a specific part in the vision structure.

\subsection{Hierarchical Structure of Images}
The study of the hierarchical structure of images has been the interest of researchers for decades, ranging from syntactic pattern recognition \cite{Fu1968SyntacticPR, You1978SyntacticSR}, graph grammars \cite{EKERS1997DefiningAP, Han2009BottomUpTopDownIP}, to and-or graphs \cite{zhu2006stochastic,Wu2010ANS,Tu2013UnsupervisedSL}, capsule networks \cite{Sabour2017DynamicRB, Hinton2021HowTR}, and hierarchical shape segmentation \cite{Yu2019PartNetAR, Mo2019StructureNetHG, Yi2017LearningHS}. Grammar models are frequently used to model hierarchical relations and build structured representations.

However, the vision structure and grammar for these works are mostly pre-defined or learned with supervision.
Previous works attempt to induce image grammar in an unsupervised manner \cite{Tu2013UnsupervisedSL, Si2013LearningAT}, but also use dense pre-defined operations on the nodes. In this paper, we propose joint grammar induction of image and language via compound PCFGs \cite{Kim2019CompoundPC} in a self-supervised manner, which eliminates most pre-defined structures.

\subsection{Grounded Vision and Language Learning}
In recent years, there have been lots of efforts and advances on exploiting the cross-modality alignment between vision and language for various tasks, such as image-text retrieval \cite{kiros2014unifying,karpathy2014deep}, image captioning \cite{karpathy2015deep,ma2015multimodal,xu2015show}, and visual question answering \cite{lu2016hierarchical,anderson2018bottom}. These works align the objects in images and the words in sentences either explicitly by building the visual-word mapping \cite{kiros2014unifying,karpathy2014deep,karpathy2015deep}, or implicitly by modeling the cross-modality attention \cite{xu2015show,lu2016hierarchical,anderson2018bottom}. Most recently, there has been a surge of interest in multi-modal pre-training for representation learning in vision-and-language tasks \cite{sun2019videobert,su2019vl,lu2019vilbert,tan2019lxmert,chen2020uniter}. These works extend BERT \cite{devlin2018bert}, a popular pre-training framework for natural language understanding, to multi-modalities by pre-training on large-scale image/video and text pairs, then fine-tuning on downstream tasks. These multimodal BERT's success greatly relies on encoding the alignment between words and image regions into attention flows in the Transformer architecture \cite{vaswani2017attention}.

In this work, we share a similar spirit of structurally aligning visual and textual elements to facilitate grammar induction in both vision and language. The intuition behind this practice is that forcing the multimodal alignment can reduce the inherent ambiguity of grammar induction for individual modalities, and the induced grammar can be more effective for downstream tasks with its structured representation.

%% file: 3_dataset.tex
\section{The \textsc{PartIt} Dataset}
We present \textsc{PartIt}, a large-scale dataset of manually annotated sentences that describe both the object-level and the part-level features of an object. To the best of our knowledge, it is the first dataset with annotated natural language sentences that describe both object semantics and fine-grained part semantics paired with images.

We use \acs{amt} to collect such sentences. Given an image of an object together with the images of highlighted parts of the object, a worker is asked to use one sentence to describe all parts of the object. The workers can describe the shape, size, and amount of the parts as well as the type of the object (\eg, a chair can be a folding chair, office chair, sofa, \etc). The annotating interface, detailed instructions, and examples that we provide for workers can be found in the \textit{supplementary material}.

We obtain $\sim$10,000 3D CAD models and their part annotations from the PartNet dataset~\cite{mo2019partnet}. We choose four categories of objects: \texttt{chair}, \texttt{table}, \texttt{bed}, and \texttt{bag}. These categories are picked because they are geometrically complex, highly diverse, and have rich grammar hierarchies. While the PartNet dataset provides part annotations at multiple levels (coarse, middle, and fine-grained) based on and-or grammar, we propose to learn the grammar in an unsupervised manner without annotation. We only take the fine-grained parts from the PartNet dataset, following their original order of decomposition. We combine certain minuscule parts (\eg, knob and connector) with their parents for simplicity. Based on the and-or templates provided by PartNet, we generate ground-truth grammar rules of each object category for evaluation only, which are listed in the \textit{supplementary material}.

\begin{figure}
    \centering
    \includegraphics[width=0.9\linewidth]{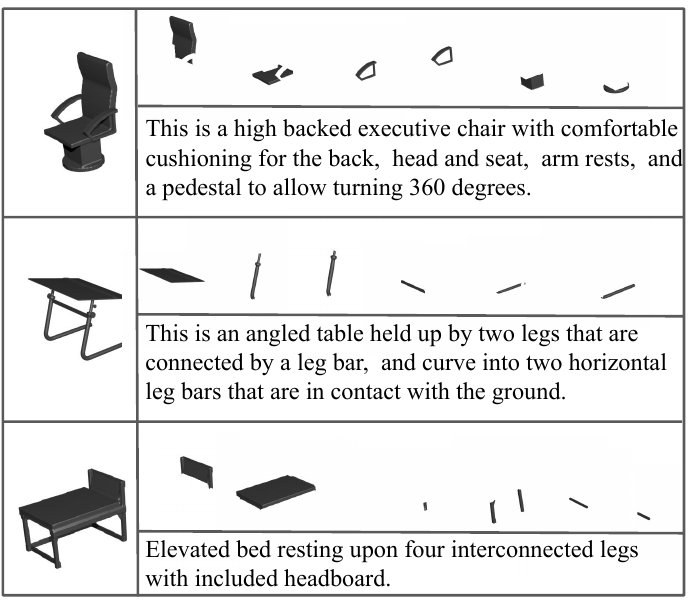}
    \caption{\textbf{Examples from our \textsc{PartIt} dataset.} The annotator is asked to use one sentence to describe all the parts of an object.}
    \label{fig:dataset}
\end{figure}

\cref{tab:dataset} shows the statistics of our dataset. We observe that the median number of grammar rules used per object is 8, which suggests that the part grammar is complex enough to be learned. For language, the median length of the sentences is 16, which is much longer comparing to existing image captioning datasets (\eg, previous visually-grounded grammar induction models~\cite{Shi2019VisuallyGN, Zhao2020VisuallyGC} use MSCOCO, which only has an average length of 10 words per sentence). Apart from grammar induction, the dataset can be used in related downstream tasks, \eg, image captioning, language-guided part segmentation, 3D reconstruction and so on. Examples of the \textsc{PartIt} dataset are shown in \cref{fig:dataset}.
\begin{table}[htbp]
    \centering
    \caption{\textbf{The statistics of the \textsc{PartIt} dataset.} \#PS is the number of part semantics, and \#G is the number of grammar rules. $P_{med}$ and $P_{max}$ denote the median and maximum numbers of part instances per object, respectively. $G_{med}$ and $G_{max}$ denote the median and maximum number of grammar rules used per object, respectively. $LG_{med}$ and $LG_{max}$ denote the median and maximum length of sentences, respectively, and $Vocab$ denote the size of the language vocabulary.}
    \label{tab:dataset}
    \small
    \begin{tabular}{c|ccccc}
        \toprule
        & \bf{All} & \bf{Chair} & \bf{Table} & \bf{Bed} & \bf{Bag} \\ \hline
        \#$I$ &10613 & 5031 & 5290 & 185 & 109\\ \hline
        \#$PS$ &120110&13 & 10&8 &3\\
        \#$G$ &75&23 & 34&18 &4\\ \hline
        $P_{med}$ &7&8 &6&9 &3\\
        $P_{max}$ &136&38 &136&28 &6\\ \hline
        $G_{med}$ &8&8 &8 &7 &3 \\
        $G_{max}$ &18&12 &18&15 &4\\ \hline
        $LG_{med}$ &16&19 &13&19 &15\\
        $LG_{max}$ &98&98 &68&42 &21\\
        $Vocab$ &2007&1634 &903&176 &61 \\
        \bottomrule
    \end{tabular}
    
\end{table}

%% file: 4_method.tex
\section{Grounded Grammar Induction}
\begin{figure*}
    \centering
    \includegraphics[width=\textwidth]{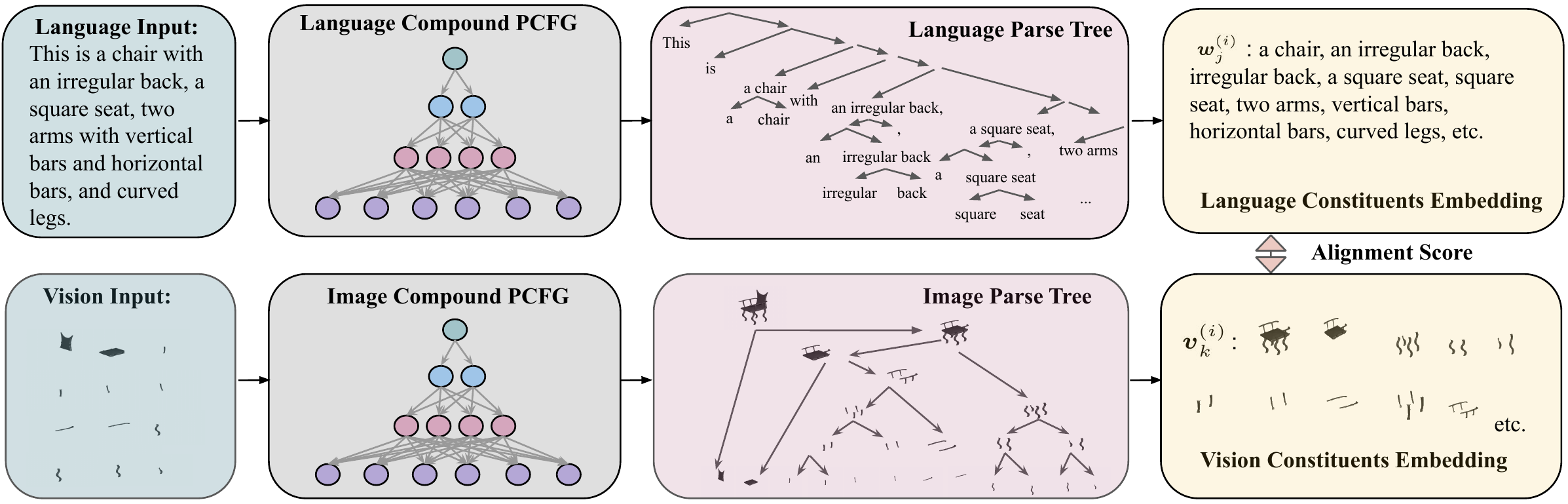}
    \caption{ \textbf{Our proposed VLGrammar framework.} We implement image grammar induction and language grammar induction via compound PCFGs. Parse trees are derived from the grammars. We compute alignment scores between the vision and language constituents in the parse trees to guide the joint learning procedure.}
    \label{fig:framework}
\end{figure*}
In this section, we introduce the proposed VLGrammar for grounded grammar induction in both vision and language. Our model starts from the compound PCFG for inducing the language grammar~\cite{Kim2019CompoundPC} and generalizes this idea to vision, which are jointly optimized by a contrastive loss.

\subsection{Compound PCFG for Language}
A context-free grammar (CFG) can be defined as a 5-tuple $\mathcal{G}=(S, \mathcal{N}, \mathcal{P}, \Sigma, \mathcal{R})$, where $S$ is the start symbol, $\mathcal{N}$ is a finite set of nonterminal nodes, $\mathcal{P}$ is a finite set of preterminal nodes, $\Sigma$ is a finite set of terminal nodes, and $\mathcal{R}$ is a set of production rules in the Chomsky normal form:
\begin{equation}
\begin{array}{rrr}
S \rightarrow A, & A \in \mathcal{N} \\
A \rightarrow B C, & A \in \mathcal{N}, B, C \in \mathcal{N} \cup \mathcal{P} \\
T \rightarrow w, & T \in \mathcal{P}, w \in \Sigma
\end{array}
\end{equation}

In natural language, nonterminals $\mathcal{N}$ are constituent labels and preterminals $\mathcal{P}$ are part-of-speech tags. A terminal node $w$ is a word from a sentence, and $\Sigma$ is the vocabulary. During implementation, we do not have the ground truth constituent labels and part-of-speech tags. Therefore, nonterminals and preterminals are sets of nodes (or clusters) which implicitly represent their functions.

Probabilistic context-free grammars (PCFGs) extend CFGs by assigning a probability $\pi_r$ to each production rule $r \in \mathcal{R}$ such that $\sum_{r: A \rightarrow \gamma} \pi_{r}=1$, \ie, the probabilities of production rules with the same left-hand-side nonterminal sum to 1. Kim et al. \cite{Kim2019CompoundPC} propose a neural parameterization where  rule  probabilities  are  based  on  distributed representations. To mitigate the strong context-free assumption, they extend neural PCFGs to compound PCFGs by assuming that rule probabilities follow a compound probability distribution \cite{Cam1986AsymptoticMI}:
\begin{equation}
\pi_{r}=g_{r}(\mathbf{z} ; \theta), \quad \mathbf{z} \sim p(\mathbf{z})
\end{equation}
where $p(\mathbf{z})$ is a prior distribution of the latent variable $\mathbf{z}$, and rule probability $\pi_{r}$ is parameterized by $\theta$. $\pi_{r}$ takes one of the following forms:
\begin{align}
&\pi_{S \rightarrow A} =\frac{\exp \left(\mathbf{u}_{A}^{T} f_{s}\left(\left[\mathbf{w}_{S} ; \mathbf{z}\right]\right)\right)}{\sum_{A^{\prime} \in \mathcal{N}} \exp \left(\mathbf{u}_{A^{\prime}}^{T} f_{s}\left(\left[\mathbf{w}_{S} ; \mathbf{z}\right]\right)\right)} \label{eqn1}\\
&\pi_{A \rightarrow B C} =\frac{\exp \left(\mathbf{u}_{B C}^{T}\left[\mathbf{w}_{A} ; \mathbf{z}\right]\right)}{\sum_{B^{\prime}, C^{\prime} \in \mathcal{N} \cup \mathcal{P}} \exp \left(\mathbf{u}_{B^{\prime} C^{\prime}}^{T}\left[\mathbf{w}_{A} ; \mathbf{z}\right]\right)} \label{eqn2}\\
&\pi_{T \rightarrow w} =\frac{\exp \left(\mathbf{u}_{w}^{T} f_{t}\left(\left[\mathbf{w}_{T} ; \mathbf{z}\right]\right)\right)}{\sum_{w^{\prime} \in \Sigma} \exp \left(\mathbf{u}_{w^{\prime}}^{T} f_{t}\left(\left[\mathbf{w}_{T} ; \mathbf{z}\right]\right)\right)}  \label{eqn3}
\end{align}
where $\mathbf{u}$ is a parameter vector, $\mathbf{w}_N$ ($N \in\{S\} \cup \mathcal{N} \cup \mathcal{P}$) is a symbol embedding. $[\cdot ; \cdot]$ indicates vector concatenation, and $f_{s}(\cdot)$ and $f_{t}(\cdot)$ are feedforward neural networks that encode the inputs.

In the compound PCFG, the log marginal likehood $\log p_{\theta}(\boldsymbol{w})$ of the observed sentence $\boldsymbol{w}=w_{1} w_{2} \ldots w_{n}$ can be obtained by summing out the latent tree structure using  the  inside  algorithm \cite{Baker1979TrainableGF}: 
\begin{equation}
\log p_{\theta}(\boldsymbol{w})=\log \int_{\mathbf{z}} \sum_{t \in \mathcal{T}_{\mathcal{G}}(\boldsymbol{w})} p_{\theta}(t \mid \mathbf{z}) p(\mathbf{z}) d \mathbf{z}
\end{equation}
where $\mathcal{T}_{\mathcal{G}}$ consists of all parses of the sentence $\boldsymbol{w}$ under a grammar $\mathcal{G}$. Since the integral over $z$ makes this likelihood intractable, Compound PCFGs use amortized variational inference and compute the loss based on the evidence lower bound (ELBO):
\begin{equation}
\small
\begin{aligned}
&\mathcal{L}_{g}(\boldsymbol{w}; \phi, \theta)= - \operatorname{ELBO}(\boldsymbol{w} ; \phi, \theta)\\
&= - \mathbb{E}_{q_{\phi}(\mathbf{z} \mid \boldsymbol{w})}\left[\log p_{\theta}(\boldsymbol{w} \mid \mathbf{z})\right] + \operatorname{KL}\left[q_{\phi}(\mathbf{z} \mid \boldsymbol{w}) \| p(\mathbf{z})\right]
\end{aligned}
\end{equation}
\noindent where ${q_{\phi}(\mathbf{z} \mid \boldsymbol{w})}$ is a variational posterior modeled by a neural network parameterized by $\phi$.

\subsection{Compound PCFG for Imgae}
Compound PCFGs can be naturally extended to image grammar. In a compound PCFG for image, $S$ denotes an object, \eg, a chair. Nonterminals $\mathcal{N}$ are types of middle-level coarse parts. Preterminals $\mathcal{P}$ are types of fine-grained leaf-parts. The middle-level parts can be further decomposed into sub-parts which are either middle-level parts or leaf-parts; for example, the base of a chair is decomposed into the central support and the leg system, and the leg system is further decomposed into several legs.

\cref{eqn1} and \cref{eqn2} can be directly applied to represent the compound PCFG for image. However, \cref{eqn3} does not work for image, since we do not have a fixed vocabulary for images, and terminal nodes are varied \textit{w.r.t} pixels. To address this problem, we design a bottom-up perception module to substitute the top-down generation in \cref{eqn3}.  

\subsubsection{Bottom-Up Perception}
Preterminals $T$ can be viewed as a set of clusters that group the terminal nodes since we do not have ground-truth labels. Therefore, instead of inducing the top-down grammar, we use a bottom-up perception module to propose terminal nodes for $T$. 

We consider the terminal nodes to be a sequence of leaf-parts of an object $\boldsymbol{v} = v_{1}v_{2}...v_{n}$. We want to assign a tag $T$ to each leaf part $v_i$.
\begin{equation}
    s(T,v_i)=\mathbf{u}_{T}^{T}f_{t}\left(\psi(v_i)\right)
    \label{eqn:perception}
\end{equation}
where $\psi$ is a perception module, \ie, ResNet-18 in our model. $f_{t}$ is a clustering model, which is a single-layer feed-forward neural network that gives the score of clustering leaf-part $v_i$ to the tag $T$ and $\mathbf{u}_T$ is a parameter vector for the tag $T$.
The rule probability of a preterminal to a leaf-part is thus:
 \begin{equation}
    \pi_{T \rightarrow v_i} = \frac{exp(s(T, v_i))}{\sum_{v^\prime \in \Sigma} exp(s(T, v^\prime))}
\end{equation} 
All leaf parts in a training batch constitute $\Sigma$.

We maximize the log-likelihood of the part sequence with ELBO:
\begin{equation}
\small
\begin{aligned}
\mathcal{L}_{g}(\boldsymbol{v} ; \phi, \theta)
= -\mathbb{E}_{q_{\phi}(\mathbf{z} \mid \boldsymbol{v})}\left[\log p_{\theta}(\boldsymbol{v} \mid \mathbf{z})\right] - \operatorname{KL}\left[q_{\phi}(\mathbf{z} \mid \boldsymbol{v}) \| p(\mathbf{z})\right]
\end{aligned}
\end{equation}
\noindent where ${q_{\phi}(\mathbf{z} \mid \boldsymbol{v})}$ is a variational posterior.

Note that the image sequence $\boldsymbol{v}$ is independent of $z$ given the tags $\boldsymbol{T}=T_1T_2...T_n$ of $\boldsymbol{v}$. Therefore,
\begin{equation}
\begin{aligned}
    p_{\theta}(\boldsymbol{v} \mid \mathbf{z}) &= \sum_{\boldsymbol{T}} p_{\theta_\psi}(\boldsymbol{v}|\boldsymbol{T})p_{\theta_\mathcal{G}}(\boldsymbol{T}|\mathbf{z})\\
    &\propto \sum_{\boldsymbol{T}}p_{\theta_\psi}(\boldsymbol{T}|\boldsymbol{v})p_{\theta_\mathcal{G}}(\boldsymbol{T}|\mathbf{z})
    \label{eqn:cluster}
\end{aligned} 
\end{equation}
where we sum over all possible tags for the part. $\theta_\psi$ denotes the parameters of the clustering module, and $\theta_\mathcal{G}$ denotes the parameters of \cref{eqn1} and \cref{eqn2} in the image grammar. 

We notice that if $\boldsymbol{T}$ has higher probability given by the grammar module, $p_{\theta_\mathcal{G}}(\boldsymbol{T}|\boldsymbol{z}) $ has a larger value, thus gives larger weight for $p_{\theta_\psi}(\boldsymbol{T}|\boldsymbol{v})$. This means $\boldsymbol{T}$ is more likely to be the accurate clusters over the images if it conforms to the current grammar. Therefore, the grammar module can boost the training of the clustering module, and \textit{vice versa}. This is demonstrated in \cref{sec:clustering}. In practice, a pre-trained clustering module can speed up the training.
 
\subsection{Joint Learning by Alignment}
We propose to jointly learn the grammars for image and language by aligning the paired image and sentence. Similar to \cite{Shi2019VisuallyGN} and \cite{Zhao2020VisuallyGC}, we use an end-to-end contrastive learning framework. While they align each language constituent with a single image, we compute an alignment score between each language constituent and each visual constituent. 
 
Given a sentence $\boldsymbol{w} = w_{1} \ldots w_{m}$ where $m$ is the total number of words, a language constituent is defined as a span over this sentence, denoted as $\textbf{w}_j = w_{a} \ldots w_{b} \in [\boldsymbol{w}]$ where $0<a<b \leq m$ and $[\boldsymbol{w}]$ denotes the set of all possible spans over $\boldsymbol{w}$. We use a Bi-LSTM to obtain the embedding of a language constituent:
\begin{equation}
\boldsymbol{w}_{j}=f_w\left(\frac{1}{b-a+1} \sum_{l=a}^{b} \mathbf{h}_l\right)
\end{equation}
where $\mathbf{h}_l$ is the hidden state of the Bi-LSTM, and $f_w$ is an affine transformation. We average the label-specific representations like in \cite{Zhao2020VisuallyGC}.

Given an object $\boldsymbol{v} = v_{1} \ldots v_{n}$ where $n$ is the total number of parts, a visual constituent is defined as a span over this part sequence, denoted as $\textbf{v}_k = v_{c} \ldots v_{d} \in [\boldsymbol{v}]$ where $0<c<d \leq n$ and $[\boldsymbol{v}]$ denotes the set of all possible sub-parts over $\boldsymbol{v}$. We define the embedding of a visual constituent as: 
\begin{equation}
\boldsymbol{v}_{k}=f_v\left(\frac{1}{d-c+1} \sum_{l=c}^{d} \psi(v_l)\right)
\end{equation}
where $\psi$ is the perception module from \cref{eqn:perception} and $f_v$ is an affine transformation.

The alignment score between a language constituent and a visual constituent is defined as their cosine similarity:
\begin{equation}
    s(\boldsymbol{w}_j, \boldsymbol{v}_k) \triangleq cos(\boldsymbol{w}_j, \boldsymbol{v}_k)
\end{equation}
The alignment score between a sentence and an image is:
\begin{equation}
\small
\begin{aligned}
&\mathcal{S}(\boldsymbol{w}, \boldsymbol{v}) = \sum_{ \substack{{t_w} \in \mathcal{T}_{\mathcal{G}_w}(\boldsymbol{w}) \\ {t_v} \in \mathcal{T}_{\mathcal{G}_v}(\boldsymbol{v})}} p (t_{w}|\boldsymbol{w})p (t_{v}|\boldsymbol{v}) \sum_{ \substack{\boldsymbol{w}_j \in t_w \\ \boldsymbol{v}_k \in t_v}} s(\boldsymbol{w}_j, \boldsymbol{v}_k)\\
&= \sum_{ \substack{\boldsymbol{w}_j \in [\boldsymbol{w}] \\ \boldsymbol{v}_k \in [\boldsymbol{v}]}} \sum_{ \substack{{t_w} \in \mathcal{T}_{\mathcal{G}_w}(\boldsymbol{w}) \\ {t_v} \in \mathcal{T}_{\mathcal{G}_v}(\boldsymbol{v})}} \mathbbm{1}_{\{\boldsymbol{w}_j \in t_w\}} \mathbbm{1}_{\{\boldsymbol{v}_k \in t_v\}} p(t_{w}|\boldsymbol{w}) p(t_{v}|\boldsymbol{v}) s(\boldsymbol{w}_j, \boldsymbol{v}_k) \\
&=\sum_{ \substack{\boldsymbol{w}_j \in [\boldsymbol{w}] \\ \boldsymbol{v}_k \in [\boldsymbol{v}]}} p(\boldsymbol{w}_j|\boldsymbol{w};\mathcal{G}_w) p(\boldsymbol{v}_k|\boldsymbol{v};\mathcal{G}_v)s(\boldsymbol{w}_j, \boldsymbol{v}_k)
\label{eqn:similarity}
\end{aligned}
\end{equation}
where $p(\boldsymbol{w}_j|\boldsymbol{w};\mathcal{G}_w) = \sum_{{t_w} \in \mathcal{T}_{\mathcal{G}_w}(\boldsymbol{w})} \mathbbm{1}_{\{\boldsymbol{w}_j \in t_w\}} p(t_{w}|\boldsymbol{w})$ and $p(\boldsymbol{v}_k|\boldsymbol{v};\mathcal{G}_v) = \sum_{{t_v} \in \mathcal{T}_{\mathcal{G}_v}(\boldsymbol{v})} \mathbbm{1}_{\{\boldsymbol{v}_k \in t_v\}} p(t_{v}|\boldsymbol{v})$ are the conditional probabilities of a constituent given the sentence/object, marginalized over all possible parse trees under the current grammars. They can be efficiently computed with the inside algorithm and automatic differentiation \cite{Eisner2016InsideOutsideAF}.  

Given a training batch $\mathcal{D} = \{\mathcal{W}, \mathcal{V}\} = \{(\boldsymbol{w}^{(i)},\boldsymbol{v}^{(i)})\}$, the contrastive loss is defined as:
\begin{equation}
\small
\begin{aligned}
    \mathcal{L}_C(\mathcal{W},\mathcal{V}) = &
    \sum_{i, m\neq i} [\mathcal{S}(\boldsymbol{w}^{(m)}, \boldsymbol{v}^{(i)}) - \mathcal{S}(\boldsymbol{w}^{(i)}, \boldsymbol{v}^{(i)}) + \delta]_{\text{+}}\\
     + &\sum_{i, m\neq i} [\mathcal{S}(\boldsymbol{w}^{(i)}, \boldsymbol{v}^{(m)}) - \mathcal{S}(\boldsymbol{w}^{(i)}, \boldsymbol{v}^{(i)}) + \delta]_{\text{+}}
\end{aligned}
\end{equation}
\noindent where $\delta$ is a constant margin, and $[\cdot]_{\text{+}}$ denotes $max(0,\cdot)$.

The overall training loss function is then:
\begin{equation}
\small
\mathcal{L}=\lambda_w\mathcal{L}_\mathcal{G}(\mathcal{W} ; \phi_w, \theta_w)+\lambda_v\mathcal{L}_\mathcal{G}(\mathcal{V} ; \phi_v, \theta_v) + \lambda_C \mathcal{L}_C(\mathcal{W}, \mathcal{V})
\end{equation}
where $\lambda_w, \lambda_v, \lambda_C$ are hyperparameters, and $\phi_t, \theta_t, \phi_v, \theta_v$ denote the parameters of the language and visual compound PCFGs, respectively.

%% file: 5_experiment.tex
\section{Experiments and Results}
\begin{table*}[htbp]
    \centering
    \caption{\textbf{The performance of grammar induction.} ``C'' and ``I'' denote corpus-level and instance-level F1 scores, respectively. ``VLG w/o SCAN'' denotes that we do not use SCAN to pretrain the unsupervised clustering module of VLGrammar.}
    \label{tab:res}
    \resizebox{\hsize}{!}{%
    \begin{tabular}{c|cc|cc|cc|cc|cc|cc|cc|cc|cc|cc}
        \toprule
        \textbf{Model} & \multicolumn{10}{c|}{\textbf{Vision Grammar}} & \multicolumn{10}{c}{\textbf{Language Grammar}} \\
        \hline
         & \multicolumn{2}{c|}{All} & \multicolumn{2}{c|}{Chair} & \multicolumn{2}{c|}{Table} &\multicolumn{2}{c|}{Bed} & \multicolumn{2}{c|}{Bag} 
          & \multicolumn{2}{c|}{All}
          & \multicolumn{2}{c|}{Chair} & \multicolumn{2}{c|}{Table} & \multicolumn{2}{c|}{Bed} &\multicolumn{2}{c}{Bag}\\
        & C & I & C & I & C & I & C & I & C & I & C & I & C & I& C & I& C & I& C & I \\
        \hline
        Left-Branch &16.4&20.2  &9.9&11.5 &21.1&26.3&\textbf{38.8}&59.4&54.2&60.0 &16.2&17.6 & 19.2 & 19.8 &13.7&15.8&10.5&12.0 &8.4 & 8.9 \\
        Right-Branch &40.8&49.1 &42.8&48.0&39.1&50.2&12.8&20.8&81.0 & 97.5&49.2 &\textbf{53.5}&43.7& 48.6 &54.2 &\textbf{58.1}  &43.7&46.2 &68.3 &69.3\\
        \hline
        ON-LSTM & /  & / & /  & /& /& /& /& /& /& / &30.7&33.4 &32.5&34.4 &28.9&32.4&27.3&29.0&39.4&38.5\\
        L-PCFG-P & /  & /& /  & /& /& /& /& /& /& / &47.8&49.4& 41.4 & 44.9 &53.6&53.5&44.9&44.3&63.7&63.5\\
        L-PCFG & /  & /& /  & /& /& /& /& /& /& / &48.4&50.3& 42.2 & 46.2 &53.6&53.5&55.3&\textbf{55.1}&71.2 &71.4 \\
        V-PCFG &47.5&59.3& 51.6 & 59.0 &43.3&59.2&36.2&48.2&82.4 &91.3 & /  & / &/  & /& /& /& /& /& /& / \\
        L-PCFG-VG & /  & /& /  & /& /& /& /& /& /& / &49.0&49.6& 42.3 & 44.0 &\textbf{54.6}&54.3&56.0 &54.6&73.0&73.0\\ 
        V-PCFG-LG &44.2&52.7& 42.0 & 47.5 &45.6&56.6&\textbf{38.8}&54.3&88.2&95.7 & /  & /& /  & /& /& /& /& /& /& / \\
        \hline
        VLGrammar &\textbf{51.4}&\textbf{63.4}& \textbf{56.4} & \textbf{65.9}  &46.3&60.5&38.1&\textbf{59.7} &\textbf{94.1} &\textbf{98.0} &\textbf{51.3} &51.9&\textbf{47.8} & \textbf{49.4} &54.0&53.8&\textbf{56.2} &54.8 &\textbf{73.6} & \textbf{73.6} \\
        VLG w/o SCAN &44.7&55.5&30.5 &33.6  &\textbf{57.9}&\textbf{75.4}&29.0&56.4& 88.2&95.7 &49.0&49.8 &43.4 &45.3 &53.7&53.5&55.1&54.0&72.6 & 72.6 \\
        \bottomrule
    \end{tabular}
    }
    
\end{table*}
\subsection{Experimental Setup}
\subsubsection{Dataset} We evaluate our model and the baseline models on the \textsc{PartIt} dataset that we collected. We obtain 2D images of the 3D objects via Blender\footnote{https://www.blender.org/}. If a part is occluded by other parts, the corresponding part image shows only the visible portion of the part. The final dataset is randomly divided into a training set of size 8,459 and a test set of size 2,154 (\ie, approximately 80\%/20\% split).

\subsubsection{Evaluation Tasks}
\textbf{Grammar Induction} We evaluate the learned grammar of both image and language. For image, we manually parse the parts into parse trees based on production rules as ground-truth. For language, we apply Benepar \footnote{https://pypi.org/project/benepar} to obtain constituency parse trees as ground-truth. We report both the averaged corpus-level F1 score and the averaged instance-level\footnote{Corpus-level F1 calculates precision/recall at the corpus level to obtain F1, while instance-level F1 calculates F1 for each visual or language instance and averages across the corpus.} F1 score against these ground-truth parse trees.

\noindent \textbf{Part Clustering} We report the accuracy of the unsupervised part clustering module to examine whether the learned image grammar can improve the part clustering results.

\noindent \textbf{Image-Text Retrieval} We evaluate text-to-image retrieval and image-to-text retrieval. When presented a sentence and eight candidate images, a model chooses the image that has the highest alignment score with the given sentence, as defined by \cref{eqn:similarity}. The image-to-text retrieval is performed likewise. 

\subsubsection{Baselines}
We compare the proposed VLGrammar with the following baselines:

\noindent \textbf{Simple tree structures} We use two simple baselines: left-branching binary trees and right-branching binary trees.

\noindent \textbf{Ordered neurons (ON-LSTM)} Shen et al. \cite{Shen2019OrderedNI} use ON-LSTM cells to predict the syntactic distance between adjacent words to induce tree structures. 

\noindent \textbf{Compound PCFGs} We use a language compound PCFG (L-PCFG) and a vision compound PCFG (V-PCFG) to induce language grammar and image grammar separately.

\noindent \textbf{Grounded compound PCFGs} Zhao et al. \cite{Zhao2020VisuallyGC} propose to learn language compound PCFGs grounded in pretrained image features, denoted as L-PCFG-VG here. For a fair comparison, we take the average of all part embeddings as the feature of an image in our setting. Similarly, we train a language-grounded compound PCFG for vision which learns image grammar grounded in pretrained language features, denoted as V-PCFG-LG.

\subsubsection{Implementation Details}
For all models that induce image grammar, we use ResNet-18 to extract the features of part images. We pretrain the ResNet-18 over unlabeled part images using an unsupervised clustering method SCAN~\cite{Gansbeke2020SCANLT}. The part features are also used in L-PCFG-VG to ground the language grammar. For V-PCFG-LG, we use BERT~\cite{Devlin2019BERTPO} for pretrained language embedding to ground the image grammar. 

Since the sentences describing different object categories share similar features in the language, we pretrain a category-agnostic language compound PCFG on the sentences across all types for 100 epochs, denoted as L-PCFG-P. Then we fine-tune the language grammar on each object category with L-PCFG, L-PCFG-VG and VLGrammar. All the models are trained for 100 epochs.
The training hyperparameters are specified in the \textit{supplementary material}. 


\subsection{Results}
\subsubsection{Grammar Induction}
\cref{tab:res} shows the main results of grammar induction of vision and language. Our method outperforms all baselines by a large margin with regard to image F1 scores. Notably, for the image grammar on \texttt{table}, VLGrammar w/o SCAN for unsupervised clustering outperforms other models significantly. It shows that our proposed VLGrammar can learn unsupervised clustering and image grammar jointly from scratch. For a category that has simple structure like \texttt{bag}, VLGrammar can achieve nearly perfect performance.  

For language grammar induction, our method is superior to all neural baselines, but slightly worse than right branching binary trees with regard to instance-level F1. The reason is that our dataset contains very long sentences, and humans tend to make right-branching sentences when the sentences are long. Similarly, the right-branching model is also a strong baseline in previous works on language grammar induction \cite{Shen2018NeuralLM, Kim2019CompoundPC}. The category-agnostic language compound PCFG (L-PCFG-P) obtains decent performance and fine-tuning it on each object category can further improves the F1 scores. One possible explanation is that while describing different objects, humans tend to use different language structures. 

\subsubsection{Part Clustering} \label{sec:clustering}
\cref{tab:cluster} shows the accuracy of the unsupervised part clustering in the bottom-up module of the image compound PCFG. Overall, after training VLGrammar, the accuracy of the part label prediction boosts from 41.3\% to 69.1\%. This confirms the argument derived from \cref{eqn:cluster}, that the induced grammar can benefit the part clustering in a top-down manner. 

One surprising observation is that even without the SCAN pretraining, VLGrammar performs quite well in the part clustering. For the \texttt{table} category, VLGrammar w/o SCAN achieves even higher accuracy than VLGrammar. The overall clustering accuracy of VLGrammar w/o SCAN is 64.4\%, which also outperforms the accuracy of SCAN (41.3\%) significantly. This can be an inspiration for unsupervised clustering: while we do not have ground truth labels, modeling the underlying structure might provide a strong learning signal for boosting the clustering.

\begin{table}[htbp]
    \centering
    \caption{\textbf{The accuracy of the unsupervised part clustering.}}
    \label{tab:cluster}
    \small
    \begin{tabular}{c|ccccc}
        \toprule
        \textbf{Model} &\textbf{All} & \textbf{Chair} & \textbf{Table} & \textbf{Bed} & \textbf{Bag}\\
        \hline
        SCAN & 41.3 & 43.5 &37.5&59.3&88.9 \\
        \hline
        V-PCFG & 61.6& 68.3 &58.3&69.9& 88.9\\
        V-PCFG-LG &65.4& 66.8 &63.2&71.8& \textbf{90.5}\\
        \hline
       VLGrammar&  \textbf{69.1} & \textbf{71.6} &66.0&\textbf{75.1}&\textbf{90.5}\\
       VLG w/o SCAN &64.4 & 62.0 &\textbf{66.2}&60.4&\textbf{90.5}\\
        \bottomrule
    \end{tabular}
    
\end{table}

\vspace{-4mm}
\subsubsection{Image-Text Retrieval}
Since an alignment score is computed to measure the similarity between an image and a sentence, it's natural to use it for image-text retrieval. 
For text-to-image retrieval, given one descriptive sentence, the model chooses the answer among eight images. For image-to-text retrieval, the model chooses among eight sentences to pair with the given object. All models are trained using contrastive loss. 
The baseline model is a simple model that uses ResNet-18 as image encoder and BERT as sentence encoder. \cref{tab:listener} shows the results. VLGrammar can outperform the baseline by a large margin and achieve satisfying performance, which is an extra bonus naturally earned with our grammar induction framework. 

\begin{table}[htbp]
    \centering
    \caption{\textbf{The accuracy of image-text retrieval.} ``IR'' stands for text-to-image retrieval and ``TR'' is for image-to-text retrieval.}
    \label{tab:listener}
    \resizebox{\hsize}{!}{%
    \begin{tabular}{c|cc|cc|cc|cc}
        \toprule
        \textbf{Model} &   \multicolumn{2}{c|}{\textbf{Chair}} & \multicolumn{2}{c|}{\textbf{Table}} & \multicolumn{2}{c|}{\textbf{Bed}} & \multicolumn{2}{c}{\textbf{Bag}}\\
        &IR & TR & IR & TR & IR & TR & IR & TR\\
        \hline
        Baseline &24.1&28.5&29.8&31.2 &20.1&20.1 &19.1 & 24.5\\
        L-PCFG-VG & \textbf{34.5} & 36.9 & 39.3 & 42.0 &35.5&\textbf{38.4} &23.0 & 28.7\\
        V-PCFG-LG & 25.9 & 27.8 & 38.8 & 41.8 &29.6&25.7&23.8 &24.9\\
        \hline
        VLGrammar & 33.2 & \textbf{39.0} & \textbf{39.8} & \textbf{42.5} &\textbf{39.6}&38.2 &\textbf{24.6} & \textbf{29.3}\\
        \bottomrule
    \end{tabular}
    }
\end{table}

\subsubsection{Cross-category Generalization}
Different object categories share certain common structures among their parts, making it possible to generalize from learned categories to unseen categories. For example, \texttt{chair}, \texttt{table} and \texttt{bed} all have legs. To evaluate the model's generalization ability, we train a shared image compound PCFG for certain object categories, and then test on unseen categories. We merge the parts and production rules of \texttt{chair} and \texttt{table}, and train a compound PCFG model on these two categories. We then test the model on all categories including two unseen categories: \texttt{bed} and \texttt{bag}. The results shown in \cref{tab:generalization} indicate that the learned grammars can indeed be transferred to novel object categories.
\begin{table}[htbp]
    \centering
    \caption{The performance of image grammars on all categories, while being trained on only \texttt{chair} and \texttt{table}.}
    \label{tab:generalization}
    \resizebox{\hsize}{!}{%
    \begin{tabular}{c|cc|cc|cc|cc}
        \toprule
        \textbf{Model} &  \multicolumn{4}{c|}{\textbf{Seen}} & \multicolumn{4}{c}{\textbf{Unseen}} \\
        \hline
         & \multicolumn{2}{c|}{Chair} & \multicolumn{2}{c|}{Table} & \multicolumn{2}{c|}{Bed} & \multicolumn{2}{c}{Bag}\\
         & C & I & C & I & C & I & C & I
         \\ \hline
         V-PCFG &43.9 &52.7 &38.1 & 54.5 & 20.7 & 33.1 & 82.4 & 91.3\\
         V-PCFG-LG &44.3 &\textbf{54.1} &38.5 &54.8 &25.6 &\textbf{50.4} &\textbf{88.2} &\textbf{95.7} \\
         \hline
         VLGrammar &\textbf{44.8} &53.4 &\textbf{41.1} & \textbf{56.7} &\textbf{29.4} & 44.2 & \textbf{88.2} & \textbf{95.7}\\
         \bottomrule
    \end{tabular}
    }
\end{table}

\vspace{-1mm}
\subsubsection{Qualitative Study}
\begin{figure}[htbp]
    \centering
    \includegraphics[width=\linewidth]{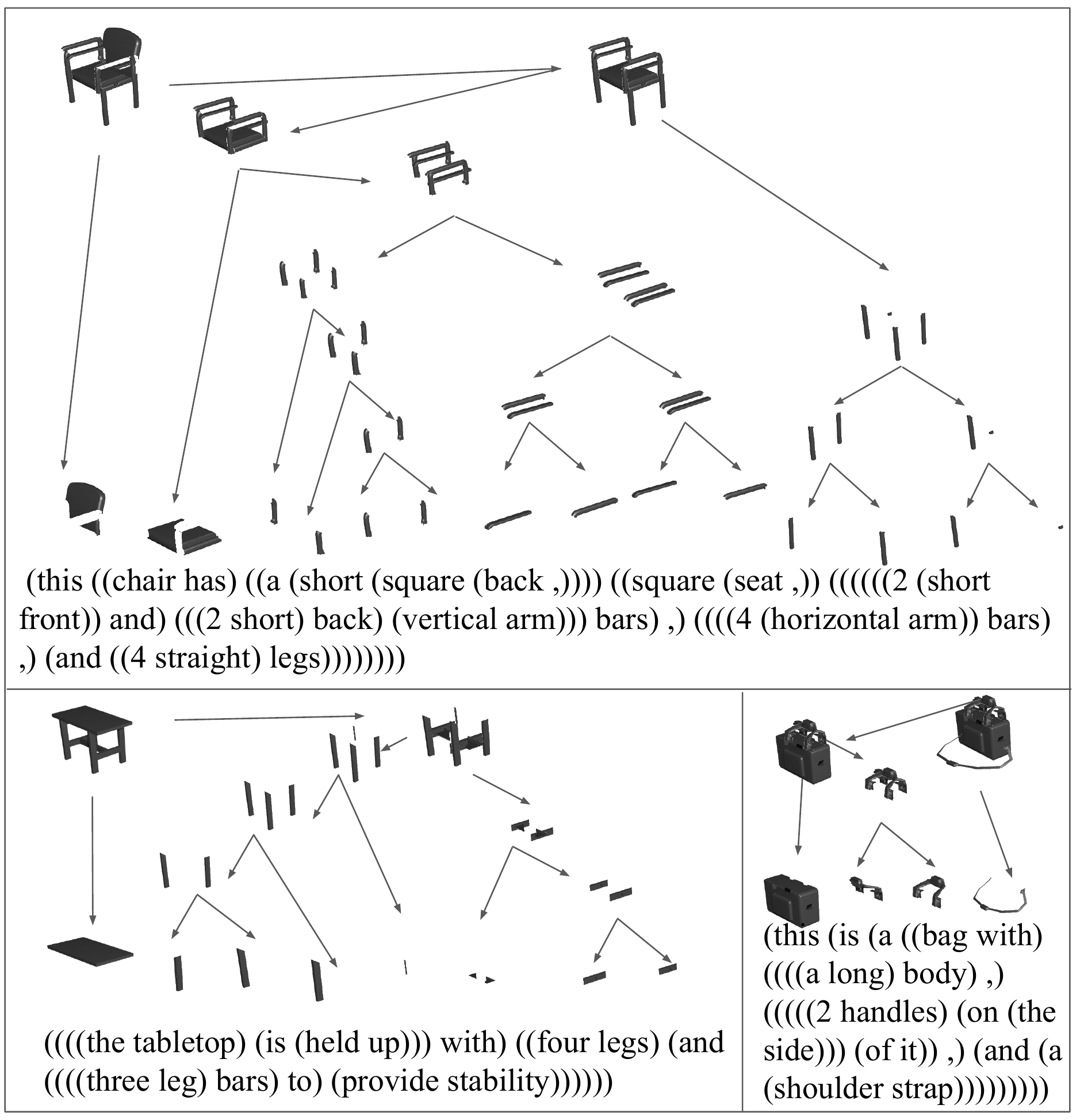}
    \caption{\textbf{Qualitative examples of parse trees predicted by VLGrammar.} We visualize the image parse trees and the language parse trees derived by the VLGrammar. Since the language parse trees are large, we use a bracket form to represent them.}
    \label{fig:qualitative}
\end{figure}
\cref{fig:qualitative} visualizes several examples of parse trees predicted by VLGrammar. We summarize following observations from these examples:

\noindent \textbf{Part-Whole Hierarchies} Our VLGrammar can capture precise part-whole hierarchies of the images. The objects can be parsed into parts of various hierarchies. For instance, the chair can be parsed into the upper part and supporting system. The latter can be further divided into seating area and chair base. The seating area has a chair seat and arms with vertical bars and horizontal bars, which are grouped separately. In the table, the base can be divided into legs and leg bars. 

\noindent \textbf{Recursive Structures}
One interesting question is how VLGrammar deals with recursive structures. A chair can have an arbitrary number of legs, which shall be in the the same hierarchy. However, since context-free grammar is defined on binary trees, recursive grammar is used to group the parts of same functionality. We find that VLGrammar can at least learn three types of recursive structures: (1) Pair-wise grouping: VLGrammar first groups the parts into pairs according to positional information (\textit{e.g.}, front legs and back legs, left horizontal arm bars and right horizontal arm bars, as shown in \cref{fig:qualitative}), and then group the pairs. (2) Right-branching: the vertical bars of the chair arms are grouped using right-branching binary trees. (3) Left-branching: \textit{e.g.}, the grouping of the legs of the table. Right-branching and left-branching are effective when dealing with an arbitrary number of parts at the same level, and when there are no salient patterns to pair them. One example is the star leg base, where the legs are in arbitrary order and form a circle.

\noindent \textbf{Language Phrases} 
VLGrammar excels at grouping phrases that refer to parts in the images. For example, VLGrammar can capture phrases such as ``a short square back", ``four legs", ``three leg bars", ``2 handles", ``a shoulder strap", and so on. This merit comes from the learned alignment between the phrases and the referred visual parts.

%% file: 6_conclusion.tex
\section{Conclusion and Future Work}
In this work, we propose VLGrammar, a framework that utilizes compound PCFGs to jointly induce the grammar of vision and language. We collect a large-scale dataset, \textsc{PartIt}, for benchmarking this novel task. Experimental results show that VLGrammar performs well in grammar induction of vision and language, greatly benefits downstream tasks such as unsupervised part clustering and image-text retrieval, and easily generalizes to unseen categories.

One limitation of our work is that the image grammar is defined on part sequences. This practice eliminates the rich 2D structures of images. A possible solution is to define spatial grammars directly on 2D images and we leave it for future work.

\section{Acknowledgements}
The  work  reported  herein  was  supported  by  ONR N00014-19-1-2153, ONR MURI N00014-16-1-2007, and DARPA XAI N66001-17-2-4029.